\pdfoutput=1

\documentclass[11pt]{article}

\usepackage[final]{acl}

\usepackage{times}
\usepackage{latexsym}

\usepackage{arydshln} 
\usepackage{booktabs} 

\usepackage{inconsolata}
\usepackage{makecell}
\usepackage{svg}
\usepackage{times}
\usepackage{latexsym}
\usepackage{graphicx}
\usepackage{tabularx}
\usepackage{xspace}
\usepackage{times}
\usepackage{booktabs}
\usepackage{latexsym}
\usepackage{graphicx}
\usepackage{multirow}
\usepackage{multicol}
\usepackage{listings}
\usepackage{amsmath}
\usepackage{makecell}
\usepackage{algorithm}
\usepackage{algpseudocode} 
\usepackage{mathtools}
\usepackage{float}
\usepackage{wrapfig}

\usepackage{pifont}
\usepackage{tcolorbox}
\usepackage{colortbl}
\usepackage{dashrule}

\usepackage[table]{xcolor}

\definecolor{my_green}{RGB}{51,102,0}
\definecolor{my_red}{RGB}{204, 0, 0}
\definecolor{api}{HTML}{ECF4FF}
\definecolor{correct_answer}{HTML}{DCF2DC}
\definecolor{CA}{RGB}{46, 151, 78}
\definecolor{CU}{RGB}{45, 125, 187}
\definecolor{NU}{RGB}{239, 120, 24}
\definecolor{SP}{RGB}{114, 97, 171}
\definecolor{MA}{RGB}{216, 37, 34}

\definecolor{gold}{RGB}{255,215,0}
\definecolor{silver}{RGB}{192,192,192}
\definecolor{bronze}{RGB}{205,127,50}

\usepackage{graphicx}
\usepackage{tikz}
\usepackage{array}
\usepackage[edges]{forest}

\usepackage{pifont}

\newcommand{\cmark}{\ding{51}} %
\newcommand{\xmark}{\ding{55}} %

\usepackage[T1]{fontenc}

\usepackage[utf8]{inputenc}

\usepackage{microtype}

\usepackage{inconsolata}

\usepackage{graphicx}

\usepackage{array}
\usepackage{wasysym}
\usepackage{amssymb}
\usepackage{marvosym}

\usepackage{multirow}
\usepackage{graphicx}

\newif\ifshowcomments
\showcommentstrue 

\title{VisualActBench: Can VLMs See and Act like a Human?}

\author{
  \textbf{Daoan Zhang\textsuperscript{\twonotes}}\thanks{Equal contribution.} 
  \quad 
  \textbf{Pai Liu\textsuperscript{\twonotes}}\footnotemark[1] 
  \quad 
  \textbf{Xiaofei Zhou\textsuperscript{\twonotes}}\footnotemark[1]
  \quad
  \textbf{Yuan Ge\textsuperscript{\Scorpio}}\footnotemark[1]
  \quad
  \textbf{Guangchen Lan\textsuperscript{\Neptune}}\footnotemark[1]
  \\
  \textbf{Jing Bi\textsuperscript{\twonotes}} 
  \quad
  \textbf{Christopher Brinton\textsuperscript{\Neptune}}
  \quad
  \textbf{Ehsan Hoque\textsuperscript{\twonotes}}
  \quad
  \textbf{Jiebo Luo\textsuperscript{\twonotes}}
  \\
  \textsuperscript{\twonotes}University of Rochester 
  \quad
  \textsuperscript{\Neptune}Purdue University
  \quad \textsuperscript{\Scorpio}Northeastern University
  \\
  \small{\texttt{daoan.zhang@rochester.edu}}
}

\begin{document}
\maketitle

\begin{abstract}
Vision-Language Models (VLMs) have achieved impressive progress in perceiving and describing visual environments. However, their ability to proactively reason and act based solely on visual inputs, without explicit textual prompts, remains underexplored. We introduce a new task, \textbf{Visual Action Reasoning}, and propose \textbf{VisualActBench}, a large-scale benchmark comprising 1,074 videos and 3,733 human-annotated actions across four real-world scenarios. Each action is labeled with an Action Prioritization Level (APL) and a proactive/reactive type to assess models' human-aligned reasoning and value sensitivity. We evaluate 29 VLMs on VisualActBench and find that while frontier models like GPT-4o demonstrate relatively strong performance, a significant gap remains compared to human-level reasoning---particularly in generating proactive, high-priority actions. Our results highlight limitations in current VLMs’ ability to interpret complex context, anticipate outcomes, and align with human decision-making frameworks. VisualActBench establishes a comprehensive foundation for assessing and improving the real-world readiness of proactive, vision-centric AI agents.
\end{abstract}

\section{Introduction}

\begin{figure} 
    \centering
    \includegraphics[width=\linewidth]{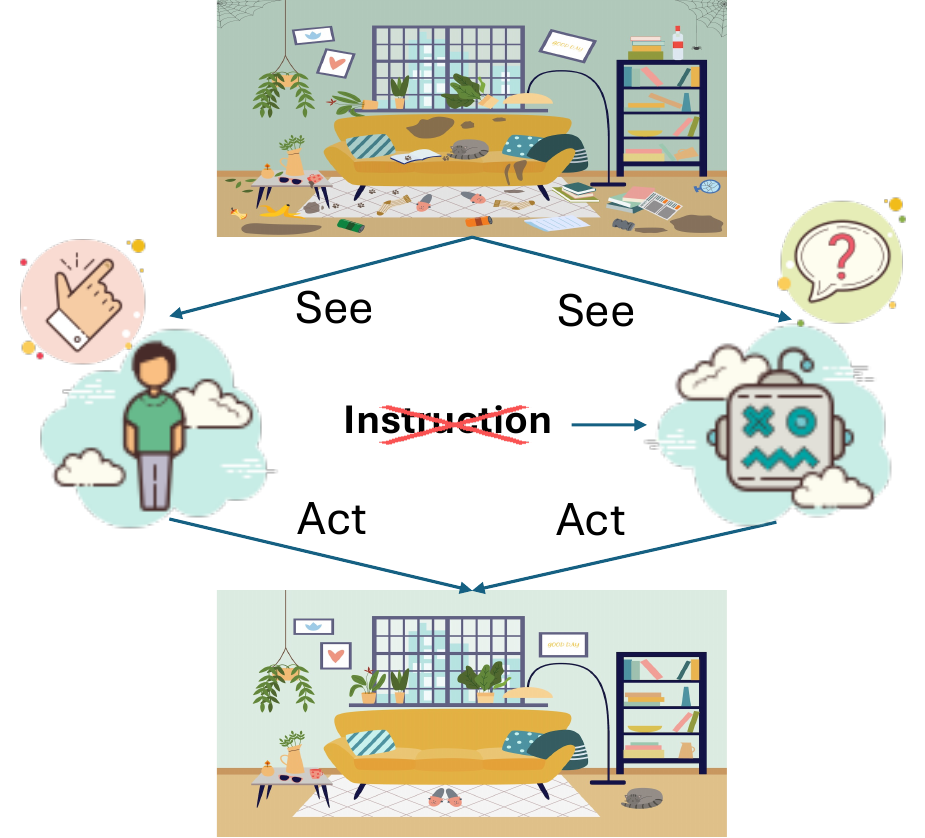}
    \caption{For humans, seeing a cluttered room naturally prompts the intention to tidy it up. However, in the absence of explicit instructions such as "tidy up the room," VLMs, relying solely on their own capabilities, infer the relevant action based on visual cues.}
    \label{fig:diagram}
\end{figure}

The emergence of Vision-Language Models (VLMs)~\cite{liu2023visual, liu2024improved, bai2023qwen, zhu2023minigpt} has greatly improved their capability to perceive and comprehend the open world. However, the advent of AGI demands a paradigm shift for VLMs: these systems must evolve from reactive language-mediated agents to vision-centric proactive entities—that is, from reliance on intermediate language representations to interpret, respond to, and act upon visual inputs, to being capable of autonomously sensing environmental dynamics through visual streams and initiating context-aware actions without linguistic intermediation.  
Comparatively, the latter approach aligns more closely with the way humans predominantly process information—through vision rather than language. For instance, humans can avoid obstacles while walking or catch a ball in mid-air by relying solely on visual perception and prior knowledge, without the need for linguistic reasoning. This vision-centered processing enables rapid decision-making and instantaneous responses, which are crucial for applications such as robotics, autonomous vehicles, and surveillance systems. However, the proactive performance of current VLMs under purely visual input conditions remains largely unknown, particularly regarding whether VLMs can generate human-consistent responses when presented with visual-only information.  

Therefore, the first research question we aim to address in this study is: \textbf{Can VLMs emulate human decision-making in open-world contexts with accurate and context-aware instructions?}

Another significant aspect of human decision-making lies in the role of behavioral value systems~\cite{schwartz2013value}. Unlike purely reactive responses, human actions are often guided by internalized values, ethical considerations, and context-specific objectives, which influence choices and prioritize outcomes beyond immediate sensory input. For instance, when crossing a busy street, a pedestrian not only processes visual cues like traffic lights and oncoming vehicles but also integrates learned behavioral norms—such as waiting for a green signal or yielding to vulnerable individuals. This value-based decision-making enables humans to navigate complex social and moral environments, ensuring their actions align with long-term goals, safety, and social acceptability.
In contrast, current VLMs primarily focus on understanding and describing visual scenes without inherently embedding or prioritizing actions based on human-aligned value systems. The models are trained to optimize performance on prompting tasks, such as object detection, image captioning, or instruction generation, often without accounting for the ethical or social implications of the responses they produce. This limitation raises an important question regarding their deployment in real-world settings, particularly in scenarios requiring nuanced decision-making, safety considerations, and adherence to social norms.
Thus, the second research question this study seeks to explore is: \textbf{Do VLMs exhibit behavior consistent with general human value systems when generating instructions in open-world contexts?}

\begin{figure*}
    \centering
    \includegraphics[width=\textwidth]{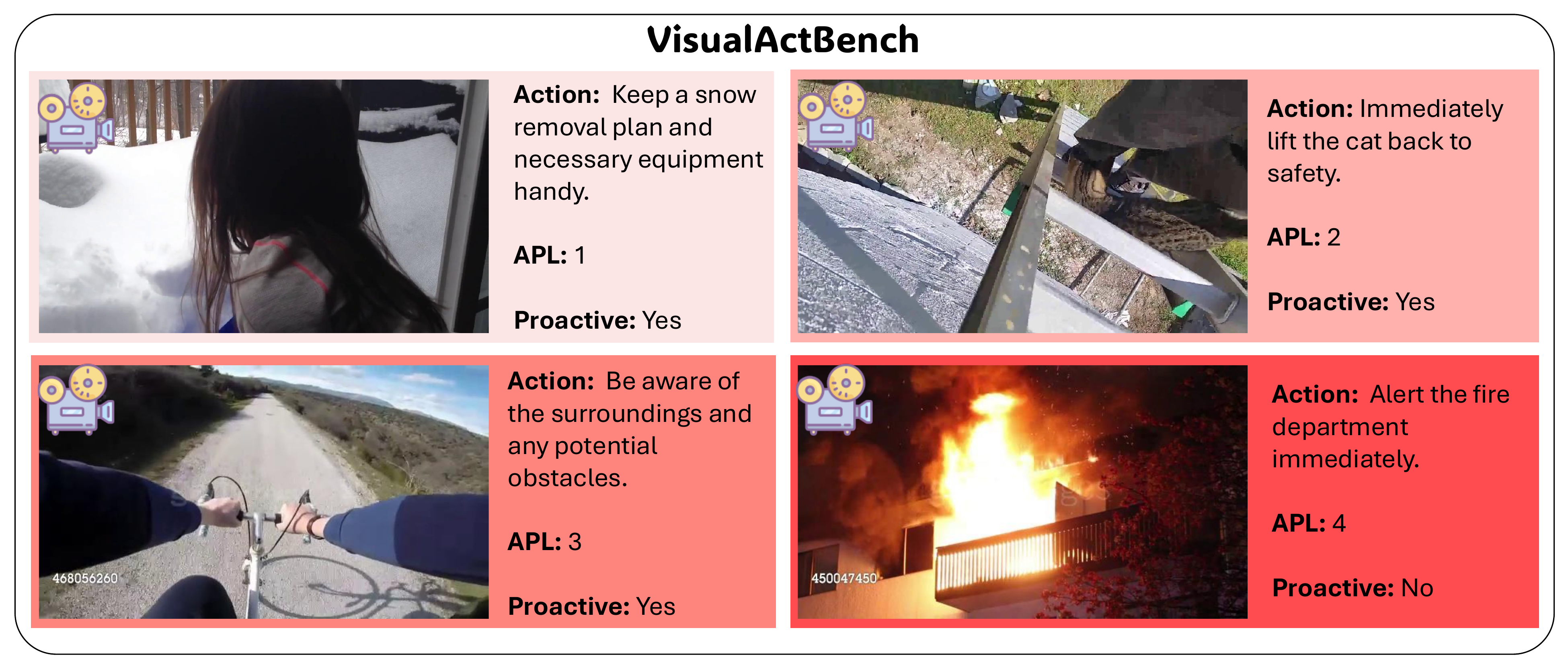}
\caption{Examples from \textbf{VisualActBench}, showcasing diverse real-world scenarios and the corresponding proactive actions with varying Action Priority Levels (APL). Each frame includes a proposed action, its associated APL, and whether the action is considered proactive.}

    \label{fig:enter-label}
\end{figure*}

To address the two key challenges outlined earlier—evaluating VLMs based solely on visual inputs and assessing their alignment with human preferences—we introduce a novel task, \textbf{Visual Action Reasoning}, along with its corresponding benchmark, \textbf{VisualActBench}. This task departs from prior work by being strictly \textit{visual-centric}, meaning that all the information required for the model to generate an appropriate response is embedded solely within the visual input, without relying on any explicit textual instructions or prompts. The goal is to assess whether VLMs can extract meaningful cues from visual scenes and reason about the most suitable actions in a manner that reflects human-like understanding and decision-making.

To construct the benchmark, we define four representative real-world scenarios: \textit{Dynamic Navigation}, \textit{Home Service}, \textit{Safety and Monitoring}, and \textit{Human-Machine Interaction}. We sample $1,074$ videos from existing datasets covering these contexts and manually annotate a total of $3,733$ actions. Each action is paired with an \textbf{Action Prioritization Level (APL)}—a metric designed to reflect human-aligned preference levels based on contextually appropriate responses. Additionally, actions are categorized into either \textit{proactive} or \textit{reactive} types, providing a lens to assess the model’s tendency toward initiative-taking behavior. This allows us to measure not only the correctness of action generation but also the extent to which models exhibit \textit{human-aligned proactiveness}.

As shown in Table~\ref{tab:bench_comp}, VisualActBench differs from traditional video benchmarks such as captioning or question-answering datasets by emphasizing action reasoning generation (ARG) and preference alignment. Instead of prompting models with textual questions or commands, our setting requires them to autonomously analyze the scene, anticipate contextually desirable outcomes, and act accordingly. This design makes VisualActBench a comprehensive and challenging benchmark that holistically evaluates both the perceptual and cognitive aspects of VLM behavior in human-like tasks.

\begin{table}
\centering
\caption{A comparative overview of various benchmarks across several dimensions, such as data format (image \textbf{I} or video \textbf{V}), the size of dataset (\textbf{\#Data)}, Visual-centric(indicated by \textbf{VC.}), Action Reasoning Generation(indicated by \textbf{ARG.}), Human Alignment(Action prioritization level, indicated by \textbf{HA.}), Open-end question(indicated by \textbf{OQ.}) and the annotation method (manual or automatic/manual, indicated by \textbf{Anno.}).}
\label{tab:bench_comp}

\resizebox{\linewidth}{!}{
\begin{tabular}{lccccccc}
\toprule
\textbf{Benchmark} & \textbf{I/V} & \textbf{\#Data} & \textbf{VC.} & \textbf{ARG.} & \textbf{HA.} & \textbf{OQ.} & \textbf{Anno.} \\ 
\midrule
\multicolumn{7}{c}{\textbf{Image Datasets (I)}} \\
\midrule
Winoground~\cite{thrush2022winoground}     & I & 400   & \xmark & \xmark& \xmark    & \xmark & M \\
MME~\cite{fu2023mme}                         & I & 1.1k  & \xmark& \xmark & \xmark    & \cmark & M \\
MMBench~\cite{liu2025mmbench}                 & I & 1.7k  & \xmark& \xmark & \xmark    & \xmark & A+M \\
MMComposition~\cite{hua2024mmcomposition}     & I & 4.3k  & \xmark& \xmark & \xmark    & \xmark & M \\
\midrule
\multicolumn{7}{c}{\textbf{Video Datasets (V)}} \\
\midrule
MSRVTT~\cite{xu2016msr}                    & V & 10k  & \cmark& \xmark & \xmark & \cmark & A+M \\
MSVD-QA~\cite{xu2017video}                    & V & 504   & \xmark& \xmark & \xmark & \xmark & A \\
MSRVTT-QA~\cite{xu2017video}                  & V & 2.9k  & \xmark& \xmark & \xmark & \xmark & A \\
TGIF-QA~\cite{jang2017tgif}                   & V & 9.6k  & \xmark& \xmark & \xmark & \xmark & A \\
TVQA~\cite{lei2018tvqa}                       & V & 2.2k  & \xmark& \xmark & \xmark & \xmark & A+M \\
ActivityNet-QA~\cite{yu2019activitynet}         & V & 5.8k  & \xmark& \xmark & \xmark & \xmark & M \\
NExT-QA~\cite{NExT-QA}                        & V & 1k    & \xmark& \xmark & \xmark & \xmark & A \\
AutoEval-Video~\cite{chen2023autoevalvideo}   & V & 327   & \xmark& \xmark & \xmark & \xmark & A+M \\
Video-Bench~\cite{ning2023video-bench}         & V & 5.9k  & \xmark& \xmark & \xmark & \xmark & A+M \\
LVBench~\cite{wang2024lvbench}                & V & 500   & \xmark& \xmark & \xmark & \xmark & M \\
MVBench~\cite{li2024mvbench}                  & V & 3.6k  & \xmark& \xmark & \xmark & \cmark & A+M \\
MovieChat-1k~\cite{song2024moviechat}         & V & 100   & \xmark& \xmark & \xmark & \cmark & M \\
TempCompass~\cite{liu2024tempcompass}         & V & 410   & \xmark& \xmark & \xmark & \xmark & M \\
Video-MME~\cite{fu2024video-mme}              & V & 900   & \xmark& \xmark & \xmark & \xmark & M \\
\midrule
HumanActBench                              & V & 1074   & \cmark& \cmark & \cmark & \cmark & A+M \\
\bottomrule
\end{tabular}
}
\end{table}

We evaluate $29$ VLMs on VisualActBench, including $24$ open-source models and $5$ proprietary models. Our findings reveal that even the best-performing VLMs exhibit a substantial gap compared to human performance on this task. Moreover, since the Visual Action Reasoning task requires both visual perception and LLM-driven reasoning capabilities, it serves as a more comprehensive evaluation of VLMs’ overall ability.

Our main contributions are:

\begin{itemize}
\item \textbf{A new vision-centric task}: We propose \textit{Visual Action Reasoning}, which challenges VLMs to generate context-aware, proactive actions directly from visual input—without any textual prompts—emulating human decision-making in dynamic environments.

\item \textbf{A human-aligned benchmark}: We introduce \textit{VisualActBench}, covering four real-world scenarios with over 3,700 annotated actions, each labeled by Action Prioritization Level (APL) and proactive/reactive type to assess value alignment and behavioral quality.

\item \textbf{A comprehensive model evaluation}: We benchmark 30 VLMs and reveal significant gaps in proactiveness, value alignment, and abstract reasoning, offering critical insights into the limitations of current models and guiding future development toward real-world deployment.
\end{itemize}

\section{VisualActBench}
\label{sec:formatting}

\begin{figure*}
    \centering
    \includegraphics[width=0.9\linewidth]{images/data_stat.pdf}
    \caption{Distribution of videos and actions in \textbf{VisualActBench}. The left two charts show the number of videos categorized by scenario type (Dynamic Navigation, Home Service, Safety and Monitoring, and Human-Machine Interaction) and by Action Priority Level (APL 0–4). The right two charts illustrate the number of actions categorized by proactivity (Proactive vs. Reactive) and by APL.}

    \label{fig:piechart}
\end{figure*}
\subsection{Video Collection}

We sampled the videos from 4 datasets, Kinetics~\cite{kay2017kinetics}, Moments in Time~\cite{monfort2019moments}, DADA-2000~\cite{fang2019dada} and UCF-Crime~\cite{sultani2018real} datasets. Notably, we select video clips that are not too long and, as much as possible, contain only a single scene. This approach helps avoid an excessive number of actions and complications related to action sequencing that may arise from multiple scenes.

\subsection{Bemchmark Statistics}
Figure~\ref{fig:piechart} provides a quantitative overview of the composition of \textbf{VisualActBench}. As shown in the leftmost pie chart, the dataset contains a balanced distribution across four key scenarios: Dynamic Navigation (364 videos), Home Service (454), Safety and Monitoring (126), and Human-Machine Interaction (130), ensuring diverse contextual coverage. The adjacent chart shows the Action Prioritization Level (APL) distribution, with the majority of videos concentrated in Level 0 (455) and Level 2 (434), while higher-risk or more critical scenarios (Level 3 and 4) are less common, reflecting the natural rarity of high-priority decision-making moments in everyday environments.

On the action side, the benchmark contains a total of 3,733 actions, as depicted in the two right-side pie charts. Notably, proactive actions significantly outnumber reactive ones (2,685 vs. 1,048), indicating that the benchmark emphasizes initiative-driven reasoning over passive response. The APL distribution of actions shows a smooth decline from Level 0 (1,675) to Level 4 (69), further confirming that while lower-priority actions are more frequent, higher-priority and more critical actions are still well-represented. This balanced distribution supports comprehensive evaluation of a model’s ability to reason across a spectrum of urgency and proactiveness.

\subsection{Evaluation Metrics}
To evaluate model performance in VisualActBench, we employ a set of metrics that capture both textual alignment and human-aligned reasoning quality. Specifically, we calculate standard precision, recall, and F1-score based on action matching, along with two additional metrics: weighted alignment score and scale score. All metrics are reported at the average level across four scenario classes.

\paragraph{Exact Matching-based Metrics.}
We use cosine similarity over sentence embeddings (produced by a SentenceTransformer model) to compute the similarity between ground-truth and predicted actions. For a pair of predicted action $p_i$ and ground-truth action $g_j$, we compute a similarity matrix $S$ where
\[
S_{i,j} = \text{cosine\_similarity}(p_i, g_j).
\]
Actions are matched using the Hungarian algorithm with a similarity threshold $\tau$ (e.g., $\tau=0.5$). Based on the matched pairs:
\begin{itemize}
    \item True Positives (TP): matched action pairs above the threshold.
    \item False Positives (FP): unmatched predicted actions.
    \item False Negatives (FN): unmatched ground-truth actions.
\end{itemize}
From these, we compute Precision, Recall and F1:

\paragraph{Weighted Matching Metrics.}
To incorporate human value alignment into the evaluation, we propose a weighted F1 variant that considers how well the predicted action reflects the importance level (scale) of the ground-truth action. For each matched pair $(p_i, g_j)$ with associated scale levels $s_i$ and $s_j$, we define a match weight as:
\[
w_{i,j} = 1 - \frac{|s_i - s_j|}{4},
\]
where the scale difference is normalized by the maximum level (4). The weighted precision and recall are then:

\begin{align}
\text{Weighted Precision} &= 
\frac{\sum_{(i,j)\in \text{matches}} w_{i,j}}{|\text{Predicted}|}, \\
\text{Weighted Recall} &= 
\frac{\sum_{(i,j)\in \text{matches}} w_{i,j}}{|\text{Ground Truth}|}.
\end{align}

and the weighted F1 score follows similarly.

\paragraph{Average Scale Score.}
To further evaluate human alignment, we introduce the average scale score, measuring the closeness of predicted action scale to ground-truth:
\[
\text{Scale Score} = \frac{1}{|\text{matches}|} \sum_{(i,j) \in \text{matches}} \left(1 - \frac{|s_i - s_j|}{4} \right).
\]
This metric penalizes mismatches in action priority even if the actions are textually similar, reinforcing the importance of value-sensitive reasoning.

For fairness across diverse scenarios, we report averaged metrics by computing the mean of each metric across the four predefined classes, which ensures equal weight for each scenario type regardless of sample size.

\begin{table*}[h!]
\centering
\caption{Comprehensive evaluation results (scores scaled to 100 and rounded to one decimal), sorted by Overall F1 (descending). In each column the highest value (top~1) is highlighted in \cellcolor{gold}{gold}, the second highest (top~2) in \cellcolor{silver}{silver} and the third highest (top~3) in \cellcolor{bronze}{bronze}.}
\label{tab:eval-results-updated}
\resizebox{\textwidth}{!}{%
\begin{tabular}{l|ccc|ccc|ccc|ccc|ccc}
\toprule
\multirow{2}{*}{\textbf{Model}} & \multicolumn{3}{c|}{\textbf{Dynamic Navigation}} & \multicolumn{3}{c|}{\textbf{Home Service}} & \multicolumn{3}{c|}{\textbf{Safety and Monitoring}} & \multicolumn{3}{c|}{\textbf{Human-Mach. Inter.}} & \multicolumn{3}{c}{\textbf{Overall}} \\
\cline{2-16}
 & Prec. & Recall & F1 & Prec. & Recall & F1 & Prec. & Recall & F1 & Prec. & Recall & F1 & Prec. & Recall & F1 \\
\midrule
GPT-4o \cite{openai2024gpt4ocard}         & 32.6 & \cellcolor{gold}{40.9} & \cellcolor{gold}{34.6}  & \cellcolor{bronze}{38.5} & \cellcolor{silver}{41.5} & \cellcolor{gold}{38.4}  & {34.5} & \cellcolor{gold}{40.9} & \cellcolor{gold}{35.8}  & \cellcolor{bronze}{37.5} & \cellcolor{gold}{42.6} & \cellcolor{gold}{38.2}  & 35.8 & \cellcolor{gold}{41.5} & \cellcolor{gold}{36.7} \\[4pt]
InternVL2.5-38B \cite{chen2025expandingperformanceboundariesopensource} & \cellcolor{silver}{40.7} & 33.9 & \cellcolor{silver}{33.9}  & \cellcolor{gold}{41.4} & 28.0 & \cellcolor{bronze}{30.9}  & \cellcolor{silver}{44.1} & 32.3 & \cellcolor{bronze}{34.6}  & \cellcolor{gold}{45.7} & 32.7 & \cellcolor{bronze}{35.3}  & \cellcolor{gold}{43.0} & 31.7 & \cellcolor{silver}{33.7} \\[4pt]
GPT4o-mini \cite{gpt4o-mini}     & 25.4 & 33.1 & 27.4  & 37.5 & \cellcolor{gold}{42.2} & \cellcolor{silver}{38.0}  & 27.6 & 34.1 & 29.2  & 35.1 & \cellcolor{silver}{41.0} & \cellcolor{silver}{36.1}  & 31.4 & \cellcolor{silver}{37.6} & \cellcolor{bronze}{32.7} \\[4pt]
Aria \cite{aria}          & 35.7 & \cellcolor{bronze}{36.2} & 32.9  & 35.0 & 30.0 & 30.3  & \cellcolor{gold}{57.2} & \cellcolor{silver}{38.2} & \cellcolor{silver}{34.7}  & 36.6 & 34.6 & 32.8  & 36.4 & 33.8 & 32.5 \\[4pt]
Gemini-1.5-Flash-8B \cite{gemini15flash8b} & 32.6 & \cellcolor{silver}{40.3} & \cellcolor{bronze}{33.7}  & 28.4 & 28.9 & {26.1}  & 32.5 & \cellcolor{bronze}{35.4} & 31.1  & {32.6} & \cellcolor{bronze}{36.1} & {31.3}  & 31.5 & \cellcolor{bronze}{35.2} & 30.6 \\[4pt]
LLaVA-OV-72B \cite{li2024llavaonevision} & 27.2 & 33.4 & 28.3  & 31.0 & \cellcolor{bronze}{31.9} & {29.9}  & 26.5 & 29.1 & 26.4  & 29.2 & 32.3 & 28.7  & 28.5 & 31.7 & 28.3 \\[4pt]
QwenVL-72B \cite{wang2024qwen2vl}   & 29.1 & 27.2 & 26.6  & 27.9 & 20.9 & 22.6  & 29.9 & 25.1 & 25.9  & 29.4 & 24.1 & 24.9  & 29.1 & 24.3 & 25.0 \\[4pt]
InternVL2.5-4B \cite{chen2024internvl25} & \cellcolor{gold}{44.7} & 22.7 & 27.4  & \cellcolor{silver}{40.2} & 16.7 & 22.0  & \cellcolor{bronze}{39.5} & 18.1 & 23.7  & \cellcolor{silver}{39.8} & 19.3 & 23.9  & \cellcolor{silver}{41.0} & 19.2 & 24.3 \\[4pt]
Gemini-1-5-Flash \cite{gemini15flash} & 25.4 & 31.1 & 25.9  & 21.6 & 21.9 & 19.6  & 23.1 & 29.7 & 24.4  & 27.2 & 27.8 & 24.4  & 24.3 & 27.6 & 23.6 \\[4pt]
QwenVL-3B  \cite{wang2024qwen2vl}   & \cellcolor{bronze}{39.5} & 21.3 & {24.2}  & 27.4 & 13.4 & 15.4  & 31.3 & 16.5 & 18.8  & 35.1 & 18.3 & 20.9  & 33.3 & 17.3 & {19.8} \\[4pt]
NVILA-8B  \cite{liu2025nvilaefficientfrontiervisual}    & 24.9 & 22.4 & 21.7  & 21.9 & 16.4 & 17.2  & 23.8 & 18.4 & 19.0  & 21.4 & 18.1 & 18.0  & 23.0 & 18.8 & 19.0 \\[4pt]
Gemini-2-Flash \cite{gemini2} & 31.6 & 20.7 & 22.9  & 28.0 & 15.3 & 18.0  & 28.8 & 19.8 & 21.5  & 29.7 & 17.3 & 20.5  & \cellcolor{bronze}{39.2} & 29.5 & 18.3 \\[4pt]
InternVL2.5-8B-MPO & 30.0 & 16.6 & 19.1  & 27.3 & 13.4 & 16.2  & 32.9 & 15.2 & 19.3  & 27.8 & 13.8 & 16.6  & 29.5 & 14.7 & 17.8 \\[4pt]
MiniCPM-o   \cite{yu2024rlaifvopensourceaifeedback}   & 30.9 & 20.0 & 22.0  & 25.4 & 14.0 & 16.8  & 29.0 & 16.6 & 20.1  & 28.8 & 16.2 & 18.9  & 38.9 & 28.5 & 16.7 \\[4pt]
InternVL2.5-8B \cite{chen2025expandingperformanceboundariesopensource}& 21.9 & 15.5 & 15.9  & 24.1 & 13.1 & 15.4  & 29.3 & 15.3 & 18.8  & 28.2 & 16.3 & 18.3  & 37.4 & 25.9 & 15.0 \\[4pt]
InternVL2.5-78B\cite{chen2025expandingperformanceboundariesopensource} & 15.4 & 15.1 & 14.4  & 18.8 & 13.7 & 15.0  & 14.9 & 11.1 & 12.3  & 20.4 & 16.9 & 17.5  & 17.4 & 14.2 & 14.8 \\[4pt]
VideoLLaMA2-7B \cite{cheng2024videollama2} & 18.4 & 18.1 & 17.4  & 14.2 & 12.4 & 12.6  & 15.9 & 15.9 & 15.1  & 13.6 & 12.1 & 12.2  & 15.5 & 14.7 & 14.3 \\[4pt]
VideoLLaMA2-7B-16F & 15.0 & 16.2 & 14.9  & 14.7 & 13.0 & 13.0  & 14.8 & 13.4 & 13.6  & 13.5 & 12.5 & 12.2  & 14.5 & 13.7 & 13.4 \\[4pt]
QwenVL-7B  \cite{Qwen2-VL}  & 32.0 & 17.7 & 20.9  & 24.1 & 11.4 & 13.9  & 29.1 & 9.7  & 12.8  & 26.1 & 13.5 & 16.1  & 29.4 & 26.3 & 13.1 \\[4pt]
NVILA-15B   \cite{liu2025nvilaefficientfrontiervisual}   & 10.5 & 12.5 & 11.2  & 8.8  & 10.7 & 9.0   & 7.9  & 9.9  & 8.3   & 12.4 & 14.2 & 12.7  & 17.2 & 9.9  & 11.8 \\[4pt]
mPLUG-Owl3-7B & 21.5 & 11.0 & 13.1  & 14.1 & 5.3 & 7.1  & 18.8 & 8.4 & 10.6  & 20.3 & 9.4 & 11.7  & 18.7 & 8.5 & 10.6 \\[4pt]
MiniCPM-v   \cite{yao2024minicpm}   & 15.7 & 9.0  & 10.7  & 12.0 & 6.8  & 8.2   & 16.5 & 8.8  & 10.9  & 10.8 & 16.2 & 10.8  & 20.5 & 14.7 & 8.6 \\[4pt]
InternVL2.5-2B \cite{chen2025expandingperformanceboundariesopensource}& 8.8  & 7.3  & 7.0   & 9.8  & 4.7  & 5.5   & 9.2  & 6.1  & 6.7   & 10.1 & 6.5  & 7.0   & 15.5 & 9.5  & 6.1 \\[4pt]
LLaVA-OV-7B \cite{li2024llavaonevision}& 7.7  & 7.1  & 7.3   & 6.0  & 4.4  & 4.8   & 3.9  & 2.7  & 2.9   & 9.8  & 9.5  & 9.4   & 6.9  & 5.9  & 6.1 \\[4pt]
VILA-1.5-40B \cite{lin2024vila} & 11.8 & 5.6  & 6.6   & 10.7 & 4.4  & 5.6   & 11.6 & 3.8  & 5.6   & 10.2 & 3.9  & 5.2   & 10.9 & 11.1 & 4.4 \\[4pt]
VideoChat-Flash-2B \cite{li2025videochatflash} & 3.3  & 1.5  & 1.9   & 6.9  & 3.1  & 4.0   & 3.3  & 1.8  & 2.2   & 7.0  & 3.1  & 4.0   & 5.1  & 2.4  & 3.0 \\[4pt]
InternVL2.5-1B \cite{chen2025expandingperformanceboundariesopensource}& 1.4  & 0.5  & 0.7   & 3.2  & 1.7  & 2.0   & 3.7  & 2.8  & 2.9   & 2.7  & 2.7  & 2.5   & 2.8  & 1.9  & 2.0 \\[4pt]
VideoChat-Flash-7B  \cite{li2025videochatflash}& 3.1  & 0.7  & 1.1   & 3.2  & 1.3  & 1.7   & 1.7  & 1.3  & 1.4   & 2.3  & 0.7  & 1.0   & 2.6  & 1.0  & 1.3 \\[4pt]
LLaVA-OV-0.5B \cite{li2024llavaonevision}& 0.0  & 0.0  & 0.0   & 0.0  & 0.0  & 0.0   & 0.0  & 0.0  & 0.0   & 0.1  & 0.1  & 0.2   & 0.0 & 0.0 & 0.1 \\[4pt]
\bottomrule
\end{tabular}%
}
\end{table*}

\section{Experiment Results}

We evaluate over 30 state-of-the-art vision-language models (VLMs) on \textbf{VisualActBench} to assess their ability to generate context-aware actions across diverse real-world scenarios. This section presents key quantitative results and insights, highlighting both overall performance and scenario-specific trends that reveal current strengths and limitations in multimodal reasoning.

\subsection{Weighted Matching Metric Comparison}
Table~\ref{tab:eval-results-updated} presents a comprehensive quantitative evaluation of 30 vision-language models (VLMs) on \textbf{VisualActBench}, focusing on their ability to generate contextually appropriate actions across four real-world scenarios—Dynamic Navigation, Home Service, Safety and Monitoring, and Human-Machine Interaction. For each model, we report precision, recall, and F1 scores scaled to [0, 100], with the top three scores per column highlighted for clarity.
Among all models, \textbf{GPT-4o} achieves the highest overall F1 score of 36.7, significantly outperforming both its smaller variant GPT-4o-mini (32.7) and other strong baselines such as InternVL2.5-38B (33.7) and Aria (32.5). GPT-4o consistently ranks in the top three across nearly all categories, confirming its advanced ability in not just visual perception but also high-level reasoning and decision-making. Interestingly, GPT-4o-mini—despite its compact size—still achieves competitive recall, particularly in Home Service and Human-Machine Interaction, indicating a strong balance between efficiency and reasoning ability.

We also observe a stark performance gap between the top-performing models and the rest of the field. Many popular open-source VLMs, including mPLUG-Owl, VideoLLaMA, and various versions of InternVL below 8B, report overall F1 scores below 20, revealing fundamental limitations in generating high-priority or proactive actions. These models often exhibit high precision but poor recall, suggesting a tendency to output conservative or generic actions, thereby missing more critical or context-sensitive responses. This imbalance indicates a failure in identifying nuanced cues that signal the need for proactive intervention, such as anticipating a fall, preventing a fire, or offering assistance.

Scenario-specific analysis further reveals that models perform best in \textit{Dynamic Navigation} and \textit{Safety and Monitoring}, where visual cues are often explicit and well-structured (e.g., obstacles, road layouts, fire). In contrast, the \textit{Home Service} and \textit{Human-Machine Interaction} domains yield lower scores across all models. These scenarios typically require the model to understand abstract intent, anticipate user needs, or consider long-term outcomes—capabilities that current VLMs largely lack. For example, generating actions like “adjust the lighting for comfort” or “check if the device is overheating” demands both commonsense and forward-looking reasoning.

Moreover, the introduction of the \textbf{Action Prioritization Level (APL)} and proactive/reactive classification in VisualActBench enables us to go beyond surface-level accuracy and examine alignment with human values. Our results show that many models default to low-APL or reactive responses, failing to reflect human preferences that favor preemptive and high-impact interventions. This aligns with our earlier findings that VLMs tend to act as passive observers rather than active decision-makers. The challenge is not only in recognizing what is happening in the scene, but in inferring what \textit{ought to be done}—a fundamentally different and cognitively demanding capability.

Qualitative inspection reveals typical failure modes such as overly generic actions (“observe the scene”), redundant outputs (“check again”), or misaligned priorities (e.g., focusing on minor details over safety-critical cues). These errors suggest that VLMs often rely on shallow correlations rather than genuine situational understanding.

\begin{table}[h!]
\centering
\caption{Scale scores (on a 0–100 scale, rounded to one decimal) for various models, sorted by Overall (from high to low). In each column, the highest value (top~1) is highlighted in \cellcolor{gold}{gold}, the second highest (top~2) in \cellcolor{silver}{silver}, and the third highest (top~3) in \cellcolor{bronze}{bronze}.}
\label{tab:scale-scores-sorted-updated}
\resizebox{\linewidth}{!}{%
\begin{tabular}{l|ccccc}
\toprule
\textbf{Model} & \textbf{DN.} & \textbf{HS.} & \textbf{SM.} & \textbf{HMI.} & \textbf{Overall} \\
\midrule\midrule
GPT-4o                & \cellcolor{gold}{69.4} & \cellcolor{gold}{64.0} & \cellcolor{gold}{70.4} & \cellcolor{gold}{61.9} & \cellcolor{gold}{66.4} \\[4pt]
GPT4o-mini            & 57.1 & \cellcolor{silver}{64.0} &  \cellcolor{bronze}{60.0} & \cellcolor{silver}{60.4} & \cellcolor{silver}{60.4} \\[4pt]
Aria                  & \cellcolor{silver}{61.6} & \cellcolor{bronze}{57.2} & \cellcolor{silver}{62.2} & \cellcolor{bronze}{59.7} & \cellcolor{bronze}{60.2} \\[4pt]
IntVL2.5-38B       & 58.1 & 55.7 & 58.3 & 59.6 & 57.9 \\[4pt]
Gemini-1.5-Flash-8B   &  \cellcolor{bronze}{59.6} & 53.8 & 59.0 & 58.7 & 57.8 \\[4pt]
LLaVA-OneVision-72B   & 53.2 & 56.8 & 52.2 & 53.7 & 54.0 \\[4pt]
Gemini-1.5-Flash      & 51.0 & 45.7 & 50.6 & 49.9 & 49.3 \\[4pt]
InternVL2.5-4B        & 52.4 & 46.2 & 45.6 & 46.3 & 47.6 \\[4pt]
QwenVL-72B            & 47.7 & 42.6 & 42.8 & 44.5 & 44.4 \\[4pt]
QwenVL-3B             & 49.4 & 34.0 & 38.4 & 40.8 & 40.6 \\[4pt]
NVILA-8B              & 47.9 & 37.0 & 38.3 & 36.9 & 40.0 \\[4pt]
Gemini-2-Flash        & 43.4 & 36.1 & 39.8 & 39.2 & 39.6 \\[4pt]
MiniCPM-o             & 41.1 & 36.3 & 36.7 & 38.9 & 38.2 \\[4pt]
InternVL2.5-8B-MPO    & 40.1 & 36.1 & 38.8 & 34.0 & 37.2 \\[4pt]
InternVL2.5-8B        & 33.1 & 31.5 & 37.8 & 37.4 & 34.9 \\[4pt]
QwenVL-7B             & 39.1 & 29.1 & 25.8 & 29.4 & 30.9 \\[4pt]
VideoLLaMA2-7B        & 33.7 & 27.7 & 31.5 & 24.6 & 29.4 \\[4pt]
VideoLLaMA2-7B-16F     & 34.0 & 29.0 & 28.5 & 25.4 & 29.2 \\[4pt]
InternVL2.5-78B       & 23.3 & 26.4 & 20.1 & 26.1 & 24.0 \\[4pt]
mPLUG-Owl3-7B         & 25.8 & 15.8 & 20.7 & 22.2 & 21.1 \\[4pt]
MiniCPM-v             & 21.6 & 17.2 & 24.5 & 20.5 & 21.0 \\[4pt]
NVILA-15B             & 16.1 & 15.4 & 12.9 & 17.2 & 15.4 \\[4pt]
InternVL2.5-2B        & 14.6 & 13.9 & 14.2 & 15.5 & 14.5 \\[4pt]
VILA-1.5-40B          & 12.9 & 12.0 & 5.6  & 12.3 & 12.0 \\[4pt]
VideoChat-Flash-2B    & 4.2  & 6.9  & 2.2  & 4.2  & 9.0  \\[4pt]
LLaVA-OneVision-7B    & 9.2  & 4.4  & 2.9  & 4.8  & 7.8  \\[4pt]
InternVL2.5-1B        & 1.9  & 3.2  & 2.0  & 5.1  & 3.6  \\[4pt]
VideoChat-Flash-7B    & 3.1  & 3.2  & 1.4  & 2.2  & 2.6  \\[4pt]
LLaVA-OneVision-0.5B  & 0.0  & 0.0  & 0.0  & 2.1  & 1.5  \\[4pt]
\bottomrule
\end{tabular}%
}
\end{table}

\subsection{Average Scale Score Evaluation}

Table~\ref{tab:scale-scores-sorted-updated} shows the normalized scale scores of different VLMs across four core scenarios, offering a more human-aligned perspective on model performance. While \textbf{GPT-4o} clearly leads with the highest overall score (66.4), models like \textbf{GPT-4o-mini}, \textbf{Aria}, and \textbf{InternVL2.5-38B} also achieve comparably strong performance, suggesting that multiple models can produce contextually appropriate actions even if they differ in architecture and size. Notably, the Safety and Monitoring (SM) scenario sees generally higher scores, likely due to its more explicit visual cues. In contrast, Human-Machine Interaction (HMI) remains challenging across the board, reflecting the difficulty of interpreting intent and social context. Some models show wide variance between scenarios, indicating specialization, while others maintain consistent but modest scores. Overall, the scale score metric highlights not only top-line performance but also qualitative stability and contextual sensitivity—key factors for real-world deployment.

\begin{figure*}[h]
    \centering
    \includegraphics[width=1\textwidth]{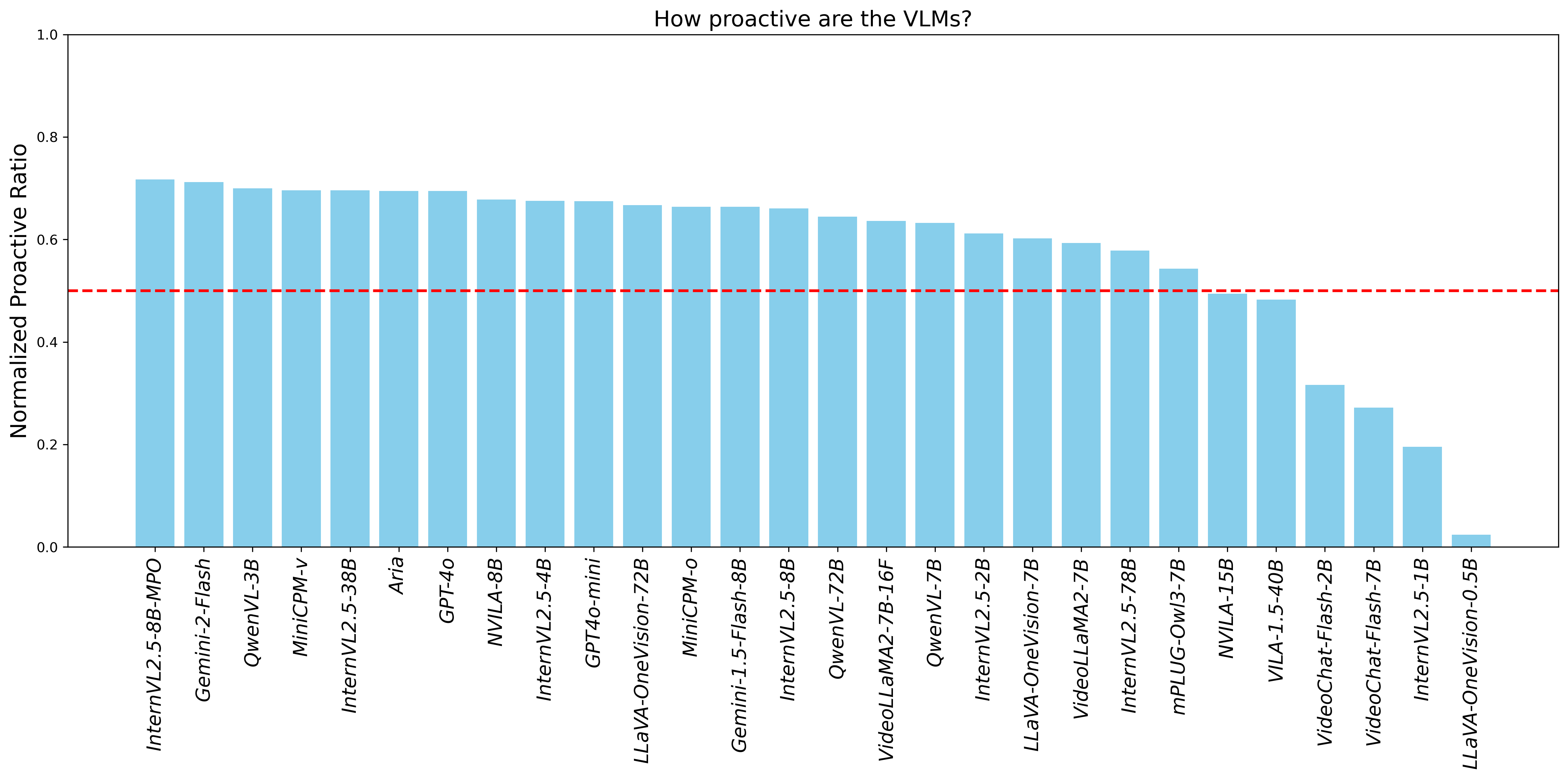}
    \caption{Normalized proactive ratios of various VLMs, reflecting their overall inclination to generate proactive rather than reactive actions across diverse scenarios. The red dashed line denotes the average proactive ratio across all evaluated models, serving as a reference for assessing the relative proactiveness and behavioral consistency of each system.}

    \label{fig:3}
\end{figure*}

\subsection{Proactiveness of VLMs}
Figure~\ref{fig:3} reports the normalized proactive ratio for each model, indicating how well each VLM completes the proactive actions expected in the evaluation set. Rather than simply preferring proactive outputs, this metric reflects a model’s ability to detect situations that require initiative and respond accordingly.

Results show that only a handful of models—such as \textbf{InternVL2.5-8B-MPO}, \textbf{Gemini-1.5-Flash}, and \textbf{QwenVL-3B}—achieve strong proactive coverage, successfully completing over 70

These findings highlight a major gap: most VLMs still struggle to recognize and act on situations where proactive, high-priority behavior is needed—limiting their applicability in safety-critical or assistance-driven environments.

\begin{table}[htbp]
    \centering
    \resizebox{\linewidth}{!}{%
    \begin{tabular}{l|c|ccc|c}
    \toprule
        Model & Frm. & Prec. & Recall & F1 & APL \\
        \midrule
        \multirow{4}{*}{LLaVA-OV} 
        & 2  & 18.6 & 8.7 & 10.6 & 17.3\\
        & 4  & 14.6 & 6.3 & 8.0 & 13.8\\
        & 8  & 6.9 & 5.9 & 6.1 & 7.8\\
        & 16 & 9.7 & 4.4 & 5.3 & 10.0\\
    \midrule
        \multirow{2}{*}{VideoLLaMA2}
        & 8  & 15.5 & 14.7 & 14.3 & 29.4\\
        & 16 & 14.5 & 13.7 & 13.4 & 29.2\\
    \bottomrule
    \end{tabular}%
    }
    \caption{Effect of varying input frame counts on model performance, showing that increasing frames from 2 to 16 does not always improve results—particularly for LLaVA-OV, where redundant visual context often reduces precision and APL, while VideoLLaMA2 maintains more stable performance across different temporal inputs.}
    \label{tab:1}
\end{table}

\begin{table}
    \centering
    \resizebox{\linewidth}{!}{%
    \begin{tabular}{l|c|ccc|c}
    \toprule
        Model & size & Prec. & Recall & F1 & APL \\
        \midrule
        \multirow{3}{*}{LLaVA-OV} 
                & 0.5B  & 0.0 & 0.0 & 0.1 & 1.5 \\
                & 7B  & 6.9 & 5.9 & 6.1 & 7.8 \\
                & 72B  & 28.5 & 31.7 & 28.3& 54.0\\
        \midrule
        \multirow{3}{*}{QwenVL} 
                & 3B  & 33.3 & 17.3 & 19.8 & 40.6\\
                & 7B  & 29.4 & 26.3 & 13.1 & 30.9\\
                & 72B  & 29.1 & 24.3 & 25.0& 44.4\\
        \midrule
        \multirow{6}{*}{InternVL} 
                & 1B  & 2.8 & 1.9 & 2.0& 3.6\\
                & 2B  & 15.5 & 9.5 & 6.1& 14.5\\
                & 4B  & 41.0 & 19.2 & 24.3& 47.6\\
                & 8B  & 37.4 & 25.9 & 15.0& 34.9\\
                & 38B  & 43.0 & 31.7 & 33.7 & 57.9\\
                & 78B  & 17.4 & 14.2 & 14.8 & 24.0\\
    \bottomrule
    \end{tabular}
    }
    \caption{Comparison of model sizes across different architectures, illustrating that larger models generally yield higher F1 and APL scores, though gains taper off at extreme scales (e.g., InternVL-78B), suggesting diminishing returns and the growing importance of data quality and training strategy beyond parameter count.}
    \label{tab:2}
\end{table}

\subsection{Impact of Frame Count and Model Size on Action Quality}
\textbf{Frame Count Analysis.}
The results are shown in Table. ~\ref{tab:1}. The number of input frames has a notable effect on model performance, but not always positively. For LLaVA-OV, increasing frame count from 2 to 16 leads to a consistent decline in F1 and APL scores. This trend can be attributed to the nature of our benchmark videos, which typically depict a single, well-defined event. Including more frames often introduces redundant information before or after the core event, thereby diluting the model’s focus and impairing its ability to generate precise and prioritized actions.
In contrast, VideoLLaMA2 demonstrates more stable performance across 8 and 16 frames, suggesting better temporal robustness. Nonetheless, the marginal differences indicate that once the core event is captured, additional frames may offer limited benefit—or even harm performance if not properly handled.

\textbf{Model Size Analysis.}
The results are shown in Table. ~\ref{tab:2}. Across all model families-LLaVA-OV, QwenVL, and InternVL—performance consistently improves with larger model size, both in terms of F1 score and APL. For instance, LLaVA-OV scales from near-zero performance at 0.5B to a strong F1 of 28.3 and APL of 54.0 at 72B. Similarly, InternVL shows a significant jump from 1B (F1: 2.0, APL: 3.6) to 38B (F1: 33.7, APL: 57.9).

However, performance gains are not always linear. InternVL-78B, for example, sees a drop in F1 and APL compared to InternVL-38B, suggesting diminishing returns or possible overfitting/misalignment at extreme scales. This hints that beyond a certain size, architectural design and training data quality may play a larger role than raw parameter count.

These findings collectively suggest that model performance in Visual Action Reasoning is constrained more by temporal abstraction and contextual grounding than by sheer model capacity. Future designs should thus focus on adaptive temporal encoding and selective frame reasoning, enabling models to extract causally relevant cues without redundancy.

\subsection{Effect of Reinforcement Learning}
As shown in ~\ref{tab:rl}, reinforcement learning (via Mixed Preference Optimization (MPO)) generally improves model performance, especially in larger models. For example, InternVL-38B shows notable gains in F1 (33.7 → 40.7) and APL (57.9 → 58.2), indicating better action quality and prioritization. While smaller models like InternVL-8B see mixed results, the trend suggests that RL enhances proactiveness and recall, helping models generate more complete and human-aligned actions.

\begin{table}
    \centering
    \resizebox{\linewidth}{!}{%
    \begin{tabular}{l|c|ccc|c}
    \toprule
        Model & RL & Prec. & Recall & F1 & APL \\
        \midrule
        \multirow{2}{*}{InternVL-4B} 
        & -  & 41.0 & 19.2& 24.3& 47.6 \\
        & MPO  & \textbf{49.3} & \textbf{20.1}& \textbf{26.5}& \textbf{48.3}\\
    \midrule
        \multirow{2}{*}{InternVL-8B} 
        & -  & \textbf{37.4} & \textbf{25.9} & 15.0 & 34.9 \\
        & MPO  & 29.5 & 14.7& \textbf{19.8} & \textbf{37.2}\\
    \midrule
        \multirow{2}{*}{InternVL-38B} 
        & -  & 43.0 & 31.7 & 33.7 & 57.9\\
        & MPO  & \textbf{49.0} & \textbf{39.9}& \textbf{40.7} & \textbf{58.2} \\
    \bottomrule
    \end{tabular}
    }
    \caption{Impact of reinforcement learning via MPO on proactive reasoning, showing consistent improvements in both F1 and APL—particularly for large-scale models—indicating that RL enhances context awareness and alignment with human-preferred action prioritization.}
    \label{tab:rl}
\end{table}

\section{Conclusion}

In this work, we presented \textbf{VisualActBench}, a novel benchmark designed to evaluate the proactive reasoning capabilities of Vision-Language Models (VLMs) in a vision-only setting. By introducing the \textit{Visual Action Reasoning} task, we challenge models to move beyond passive visual understanding and toward human-like, initiative-driven decision-making grounded in visual context alone. Through the annotation of 3,733 actions across four diverse scenarios, along with a novel Action Prioritization Level (APL) metric, VisualActBench provides a rich framework for evaluating both the correctness and value-alignment of model outputs.
Our comprehensive evaluation of 29 state-of-the-art VLMs reveals key insights: current models struggle with proactiveness, value-sensitive prioritization, and abstract reasoning. Although large-scale proprietary models such as GPT-4o perform best, even they fall short of human-level performance. Notably, increasing model size and applying reinforcement learning techniques improve action quality and prioritization, but diminishing returns are observed beyond a certain scale.
Overall, \textbf{VisualActBench} exposes critical limitations in current VLMs’ ability to act autonomously and appropriately in real-world, open-ended contexts. We hope this benchmark will serve as a stepping stone for the development of more aligned, robust, and contextually aware vision-language agents capable of operating without explicit instructions---truly seeing and acting like humans.

\bibliography{custom}

\appendix

\end{document}